\documentclass[journal]{IEEEtran}

\usepackage{cite}
\usepackage{amsmath,amssymb,amsfonts}
\usepackage{algorithm}
\usepackage{algpseudocode}
\usepackage{graphicx}
\graphicspath{{./figures/}}
\DeclareGraphicsExtensions{.pdf,.jpeg,.png}
\usepackage{xcolor}
\usepackage{units}
\usepackage{lipsum}
\usepackage[caption=false, font=footnotesize]{subfig}
\usepackage{textcomp}
\usepackage{multirow}
\usepackage{array}
\usepackage[capitalise]{cleveref}
\usepackage{url}
\usepackage{blkarray}

\let\labelindent\relax
\usepackage[inline]{enumitem}

 \usepackage{booktabs} 

\DeclareMathOperator*{\argmin}{arg\,min}

\def\BibTeX{{\rm B\kern-.05em{\sc i\kern-.025em b}\kern-.08em
    T\kern-.1667em\lower.7ex\hbox{E}\kern-.125emX}}

\newcommand{\ie}{\textit{i.e.,}\ }
\newcommand{\eg}{\textit{e.g.,}\ } 
    
\newcount\Comments  
\usepackage{color}
\definecolor{darkgreen}{rgb}{0,0.5,0}
\definecolor{purple}{rgb}{1,0,1}
\newcommand{\kibitz}[2]{\ifnum\Comments=0\textcolor{#1}{#2}\fi}

\begin{document}

\title{Towards Predicting Collective Performance in Multi-Robot Teams}

\author{Pujie Xin,~\IEEEmembership{Student Member,~IEEE,} Zhanteng Xie,~\IEEEmembership{Student Member,~IEEE,} and Philip Dames,~\IEEEmembership{Member,~IEEE} 
\thanks{This work was funded by NSF grant CNS-2143312.}
\thanks{$^{1}$P. Xin, Z. Xie and P. Dames are with  the Department of Mechanical Engineering, Temple University, Philadelphia, PA 19122, USA
        {\tt\small \{pujie.xin, zhanteng.xie, pdames\}@temple.edu}}%
}



\maketitle

\begin{abstract}
The increased deployment of multi-robot systems (MRS) in various fields has led to the need for analysis of system-level performance. However, creating consistent metrics for MRS is challenging due to the wide range of system and environmental factors, such as team size and environment size. This paper presents a new analytical framework for MRS based on dimensionless variable analysis, a mathematical technique typically used to simplify complex physical systems. This approach effectively condenses the complex parameters influencing MRS performance into a manageable set of dimensionless variables. We form dimensionless variables which encapsulate key parameters of the robot team and task. Then we use these dimensionless variables to fit a parametric model of team performance. Our model successfully identifies critical performance determinants and their interdependencies, providing insight for MRS design and optimization. The application of dimensionless variable analysis to MRS offers a promising method for MRS analysis that effectively reduces complexity, enhances comprehension of system behaviors, and informs the design and management of future MRS deployments.
\end{abstract}


\section{Introduction}
\IEEEPARstart{D}{istributed} multi-robot systems (MRS) are increasingly utilized across a broad spectrum of fields, including search and rescue \cite{Murphy2004Human-robot, Queralta2020}, surveillance \cite{Doitsidis2012}, mapping \cite{Deutsch2016SLAM}, and reconstruction \cite{Dong2019Reconstruction}. These systems are highly valued for their scalability, resilience \cite{Zhou2023}, redundancy, and efficiency. Many successful algorithms have addressed challenges in distributed MRS, such as coordinating robot movements, managing uncertainties, and optimizing search efficiency. Typically, these algorithms demonstrate strong performance in specific environments with particular sets of parameters, such as swarm size, task size, perception capabilities. This specificity is understandable, given the high-dimensional nature of the parameter space and the vast number of possible parameter combinations, which can be either continuous or discrete. However, system designers face the challenge of identifying the right set of parameters and forecasting system performance when deploying MRS algorithms in unknown environments. Consequently, there is a need for an empirical model that correlates system performance with system parameters.

Building such a model is a formidable task filled with numerous questions. Does a quantifiable relationship between these variables exist? How accurate can the predictions be? Can this model be generalized across different scenarios? These are critical inquiries that guide the development of a reliable and effective predictive model for MRS performance. Given the vast array of algorithms in distributed MRS, it is impractical to prove that any single model could be universally applicable. Therefore, we will focus our analysis on the application of Multi-Robot Multi-Target Tracking (MR-MTT), one of the most important problem in MRS, and demonstrate its effectiveness in this context.

\subsection{MR-MTT}
MR-MTT is an area where the complexity and diversity of operational scenarios are particularly pronounced \cite{robin2016multi}. MR-MTT has many potential applications, including surveillance, search and rescue, environmental monitoring, and industrial automation. In MR-MTT, a team of robots coordinate to monitor a group of targets in a given environment. We assume that the robots operate in a known area where the number of targets is initially completely unknown (\eg the robots know the blueprint map of a building but not how many people are inside). This problem is challenging due to: 
\begin{enumerate*}[label=\arabic*)]
    \item The potential mobility of targets, whose motion patterns are typically, at best, partially known.
    \item The possibility of misidentification or missed detection of targets, coupled with the uncertainty and variation in their number over time.
    \item The robots' limited access to environmental information, constrained by their sensor field of view (FoV) and communication capabilities.
\end{enumerate*}

In addressing the challenges of MR-MTT, our previous work \cite{dames2020distributed, xin2022comparing, xin2024comparison} introduced algorithms based on probabilistic search methods, which are particularly well-suited to target tracking problems given the significant noise present in the robots' sensors. Rather than providing worst-case guarantees, this approach evaluates the probabilities of target detection \cite{robin2016multi, stone2013bayesian}. This approach often involves mapping these probabilities onto the environment, creating a `heatmap' of sorts that guides the search effort. The primary reason for this methodology is the frequent lack of resources (such as robots or time) to address the worst-case scenario (ensuring the detection of a target if it is present \cite{chung2018gues}). However, it can also represent a strategic compromise between efficiency and the likelihood of encountering particularly challenging situations. Strategies may involve prioritizing specific areas based on their probability scores or dynamically allocating resources as the search progresses.

The inherent complexity of MR-MTT, arising from a vast variable space that includes the number of robots, number of targets, perception capabilities, motion models, and constraints such as communication and environment, makes it challenging to identify an optimal set of parameters for improving performance when reusing MR-MTT algorithms. Similarly, these constraints and the extensive variable space complicate the comparison and analysis of specific algorithms.

\subsection{Dimensionless Variables}
We consider to use dimensionless variables, because it offers a method to scale down the complexity and preserve the essence of the problem. And it often allows for the derivation of universal solutions that are independent of the specifics of the system \cite{Maltbaek1980, Thomas2007,Xie2022}. One example of a dimensionless variable is the Reynolds number. Originating from the study of turbulent Rayleigh–B\'enard convection, a classic problem in fluid mechanics, the Reynolds number is defined as a power-law relation of four physical quantities: fluid density, average fluid velocity, pipe diameter, and dynamic fluid viscosity. The value of the Reynolds number dictates whether the flow will be turbulent or laminar and works across a range of length scales, fluids, etc.

The concept of dimensionless variables plays an essential role in the analysis of physical systems, allowing for normalized comparisons and simplifying mathematical models. Two systems may appear different due to varying units or scales, but they may be identical or similar when expressed in dimensionless form. In our case, this can reveal fundamental characteristics of the MRS, such as the ratio of robot number to target number, which could influence system task allocation, efficiency, reliability, and cost-effectiveness. Thus, we investigate to model the MRS performance based on dimensionless variables to understand, compare and optimize MRS algorithms.  

\subsection{Approach}
To study trends in the performance of MRSs, we will utilize datasets from our previous work \cite{dames2020distributed, xin2022comparing, xin2024comparison}, which all address the problem of distributed control for MR-MTT.
We will utilize the data from these papers and run additional studies to obtain representative data that includes variations in the number of robots, the number of targets, and the FoV of the sensors.
We measure the performance of the team using two metrics.
The first is the Optimal SubPattern Assignment (OSPA) metric \cite{Schuhmacher2008Metric}, a commonly used metric in the MTT community that measures the average error of all targets.
The second is the Exploration Inefficiency (EI), which measures the exploration progress.
With either metric, we typically see a rapid initial decline that approaches an asymptote as exploration progresses.

Based on this observation, we fit two different models from the exponential family: an exponential and a sigmoid.
We note that the parameters of these models depend upon the parameters of the robot team and the task, \eg number of robots or number of targets.
However, there is not a clear trend for how changes in any single team/task parameter affect team performance.
The primary novelty in our work is the creation of a dimensionless variable that effectively describes these trends in the exponential function parameters as team/task parameters vary.
More specifically, we utilize the Buckingham Pi Theorem from the high-dimensional variable space and co-optimize the structure of the dimensionless variable and its relationship to the parameters in the exponential models.
This allows us to interpolate and extrapolate between existing samples to accurately predict team performance in novel scenarios.
It also provides insight into the team/task parameters that most affect team performance, allowing practitioners to create teams to achieve certain levels of performance.
Finally, we demonstrate that the structure of the dimensionless variable is consistent across exponential models, performance metrics, and control algorithms, indicating that the parameter provides insight into a broad range of scenarios.


\subsection{Contributions}
The key contributions of this paper are:
\begin{enumerate*}[label=\arabic*)]
    \item We identify key parameters $\theta$ that relate to the ability of an MRS to complete an MR-MTT task.
    \item We develop an algorithmic pipeline that co-optimizes the structure of a dimensionless variable $\Pi(\theta)$ (as a function of the parameters $\theta$) and the relationship between $\Pi$ and key indicators of MR-MTT success.
    \item We demonstrate that our pipeline is able to accurate predict MR-MTT performance in scenarios outside of the training dataset.
    \item We discover a dimensionless variable $\Pi(\theta)$ that is consistent across different MR-MTT search algorithms, parametric model structures, and MR-MTT performance metrics, showing that this $\Pi(\theta)$ has strong generality.
    \item We will make our source code and the underlying dataset publicly available upon publication.
\end{enumerate*}

\section{Related Work}
\label{sec:Related Work}
The study of how MRS parameters affect system-level performance originates in evaluating the varied forms and scales of MRS, which can span orders of magnitude. The primary MRS parameter discussed in the literature is the quantity of robots involved in the system \cite{Hamann2018}. From an optimization perspective, the number of robots, which are frequently regarded as potential solutions, is tied to the scope of the exploration executed by a system. This association has been thoroughly discussed in the works of \cite{PIOTROWSKI2020100718, Kwa2020Optimal}. A modest increment in the count of robots could meet the requirements if the MRS showcases scalability in the vicinity of the optimized value. 

It is widely recognized that a larger number of robots in MRS typically yields better performance. However, many studies are also delving into the intricate correlations between MRS performance and other parameters \cite{kwa2023effect}, such as robot density in multi-robot coverage path planning problem \cite{Choset1998}. In \cite{Qingbiao2021}, the authors assessed their algorithms by considering various levels of robot density. This focus on understanding the incremental or marginal utility derived from changes in parameters other than the number of robots offers valuable insights for optimizing MRS configuration. 

Research explores the relationship between the swarm size and distributed MRS performance with a given task and environment \cite{Hamann2018, Hunt2019, SCHRANZ2021100762}. The concept of swarm density often forms the basis of these studies \cite{Hamann2022}. A consistent pattern emerges across many investigations: performance increases with system size $N$ up to a critical point $N_c$, followed by declining performance for $N > N_c$. Remarkably, diverse systems from various research domains demonstrate comparable scalability trends \cite{Ward1993, Mateo2017}. A recent study on MRS performance in single-target tracking identified a phase transition in tracking efficiency that correlates with swarm density, which itself is a function of both swarm size and environmental dimensions \cite{kwa2023effect}. Performance might exhibit a bimodal distribution, indicating two distinct phases: a high-performance phase where robots operate efficiently most of the time, and a low-performance phase characterized by significant congestion trapping most robots, with little gradation between the two \cite{kuckling2023run}. Rather than using the homogeneous robots, heterogeneous robots utilizing each robot's unique sensing and mobility capabilities shows competitive performance in achieving tasks such as exploration and patrolling \cite{Salam2023}. 

Considering the spatial distribution of robots in MRS is another intuitive step. Many work suggest that there is a minimum critical density essential for fostering effective self-organization \cite{Hannes2021CIMAX} and maintaining the k-connectivity \cite{Wenhao2019} in MRS. Concurrently, there is a maximum critical density, primarily dictated by the communication capabilities of the robots, beyond which exploration becomes inefficient, potentially leading to unnecessary resource utilization. 

One common question in these studies is how factors other than team size and density, such as robots' sensing range and velocity, affect performance. By understanding how the performance of the team is impacted by these variables, it is possible to optimize the design and implementation of the system to meet these requirements \cite{Schroeder2019}.
Moreover, while these studies focus on steady-state system performance, examining performance over time through predictive analytics offers significant benefits. By analyzing current and historical data, potential deviations from normal operations can be identified early, allowing for proactive measures to prevent issues. This predictive approach also enables dynamic task reassignment, optimizing workflows and reducing bottlenecks by anticipating task completion times.

Our approach builds on this literature.
We are interested in studying not only in the effects of changing parameters but also in predicting the performance of an MRS under specific parameter configurations. This predictive capacity can assist in the strategic planning and optimization of MRS deployments.
We believe that the use of dimensionless variables can play a crucial role in both the understanding and predictive modelling of systems, including MRS, because of their ability to describe trends across various scales or units. By identifying and using dimensionless variables, we can simplify complex systems, discerning fundamental relationships that are not necessarily apparent in the raw parameters. Therefore, their application can be particularly beneficial in the study and operation of multi-robot systems.

\section{MR-MTT Dataset}
\label{sec:dataset}
We investigate the key parameters that influence the system performance in our previous work \cite{dames2020distributed, xin2022comparing, xin2024comparison}. We run simulated experiment over different set of parameters including search algorithms, number of robots, number of targets, and sensing radius.

\subsection{MR-MTT Search Algorithms}
As previously stated, we utilize search strategies from our previous works \cite{dames2020distributed, xin2022comparing, xin2024comparison}.
The first strategy is Lloyd's algorithm, which we adapted from its use for MRS coverage control for MR-MTT in \cite{dames2020distributed}.
The other four methods come from optimization theory: simulated annealing (SA), particle swarm optimization (PSO), ant colony optimization (ACO), and artificial immune systems (AIS).
Each of the optimization methods is adapted from the traditional case of finding a single global optimum to locate all the local maxima (\ie targets) (see \cite{xin2022comparing, xin2024comparison} for more details).
In each case, the MRS uses a distributed version of the probability hypothesis density (PHD) filter \cite{Mahler2003Multitarget} first developed in \cite{dames2020distributed}.

\subsection{Key Parameters}
\begin{figure}[t]
\centering
\includegraphics[width=0.52\textwidth]{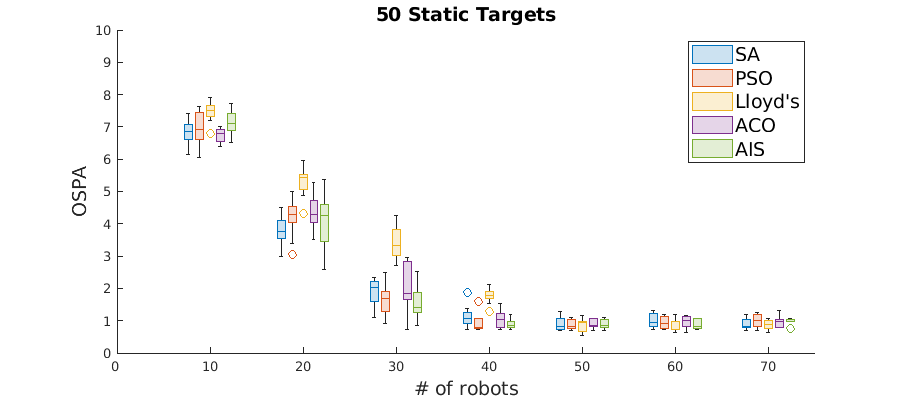}
\caption{Comparison of steady state median OSPA errors across 10 trials for various distributed control algorithms, considering a range of robot numbers and 50 static targets.}
\label{fig:50StaticTargets}
\end{figure}

Figure~\ref{fig:50StaticTargets} shows the steady-steady OSPA (\ie error) for different numbers of robots search for 50 static targets. We can see a consistent trend across all search methods: as the number of robots increases, the final OSPA value decreases, signifying enhanced tracking performance due to the larger robot team. Interestingly, once the number of robots surpasses the number of targets, the OSPA value tends to plateau, indicating minimal or no additional improvement. This plateau effect implies that, after a certain threshold, adding more robots does not substantially improve the accuracy of tracking static targets.

Predicting the trend of system performance at steady state is relatively straightforward, especially when considering single variables. For instance, it is intuitive to assume that an increase in the number of robots or the size of their FoV enhances search and tracking capabilities. Conversely, an increase in the number of targets elevates the required search capacity. However, the situation becomes more complex when multiple variables change simultaneously, particularly when the goal is to predict system performance over time, rather than at steady state. Obtaining a quantitative forecast under these conditions is both more valuable and more challenging. 

In this paper, we undertake an analysis based on five parameters across various algorithms: the number of robots ($n_r$), the number of targets ($n_t$), the radius of the FoV for each robot ($r$), the robot density in the environment ($\rho_r$), and the target density in the environment ($\rho_t$). We define $\theta$ as the parameter vector containing these variables, formally expressed as $\theta = \left[n_r, n_t, r, \rho_r, \rho_t\right]^T$. Robot density affects coverage capabilities and the frequency of interactions among robots, which in turn impacts the efficiency of task allocation and the complexity of coordination. Meanwhile, target density determines the difficulty of the tracking task; higher densities may complicate the identification and tracking of individual targets. Considering that the size of the environment may vary for robots and targets, it is necessary to account for these densities separately.


\subsection{Dataset Collection}
This study will address MR-MTT within an open \unit[{100$\times$100}]{m} region. 
All robots start at random locations within a confined box situated towards the lower central region of the map. These robots, which are holonomic, have a maximum velocity of \unit[2]{m/s}. Each robot is fitted with an onboard sensor providing a circular FoV with a radius ranging from $r=\unit[3-7]{m}$. The false negative rate in detection is set at $p_{\rm fn} = 0.2$ (\ie a robot fails to detect a target within its field of view 20\% of the time). For each target that is detected, the robot receives a noisy measurement of the relative position, drawn from a normal distribution with 0 mean and covariance $0.25I_2$. The sensors record measurements at a rate of \unit[2]{Hz}. 
We vary the number of robots in the team and targets in the search space from 10 thru 100, in increments of 10.
Targets positions are drawn uniformly at random from within the search space.
It is important to clarify that these values do not correspond to a specific real-world scenario but could resemble a drone team equipped with downward-facing cameras. For real-world applications, one could adopt the procedures detailed in our other works to accurately characterize sensor models \cite{chen2022semantic}. 

For each team/task configuration, we run 10 trials, a number we have found to be representative in our previous studies. 
This provides us with a large dataset from which we can extract trends.

\begin{figure*}[tbph]
    \centering
    \includegraphics[width=0.95\textwidth]{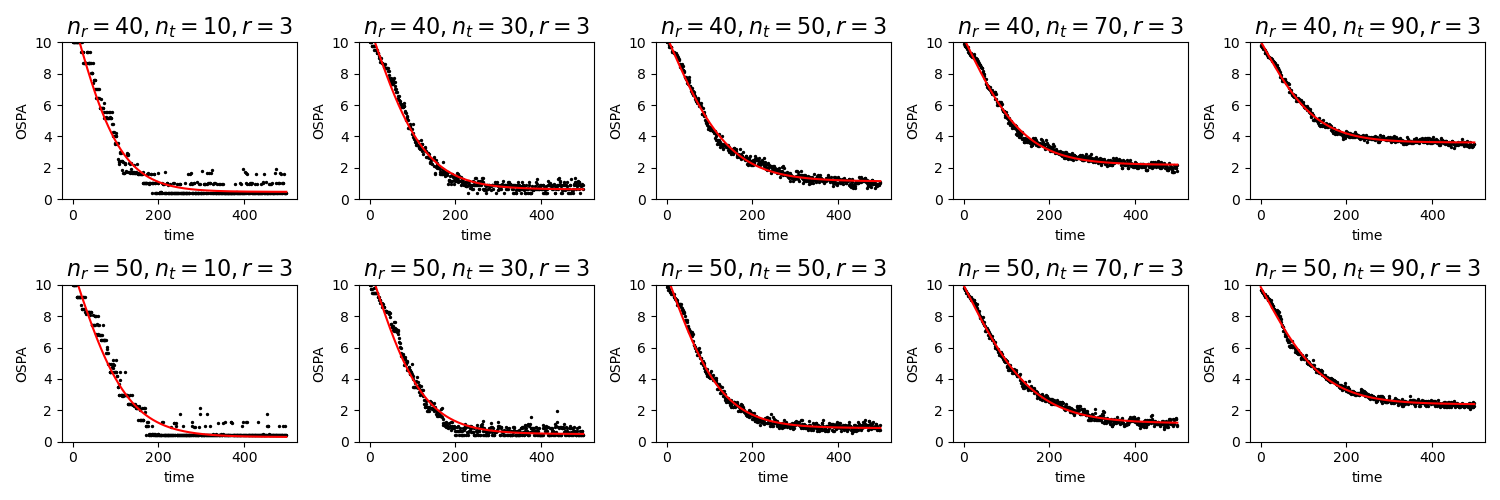}
    \caption{The black points represent the median OSPA value from the dataset with a batch size of 10 over time. The red lines illustrate the fitting results using a sigmoid function.}
    \label{fig:pureFittingresult}
\end{figure*}

\subsection{Performance Metrics}
We utilize two performance metrics in our study, one based on tracking error and the other on exploration rate.
We measure both transient and steady state performance of the MRS, where we use the median value over the final \unit[50]{s} as our indicator of the steady-state value.
Additionally, we take the median value of the metric across the 10 trials in order to decrease the impact of stochastic error and increasing the reliability of our results. Through this method, we are able to derive discrete data points, subsequently producing a comprehensive dataset. 


\subsubsection{Optimal SubPattern Assignment Metric}
The Optimal SubPattern Assignment (OSPA) metric \cite{Schuhmacher2008Metric} is a commonly used metric in the MTT community that measures the average error of all targets.
The OSPA between two sets $X,Y$, where $|X| = m \leq |Y| = n$ without loss of generality, is
\begin{equation}
    d(X,Y) = \left(\frac{1}{n}\min_{\pi \in \Pi_n}\sum_{i=1}^{m} d_c(x_i,y_{\pi(i)})^p + c^p(n-m) \right)^\frac{1}{p}
    \label{eq:OSPA}
\end{equation}
where $c$ is the cutoff distance, $d_c(x,y)=\min(c,\|x-y\|)$, and $\Pi_n$ is the set of permutations of $\{1,2,3,...,n\}$. We set $p=1$ as the $p$-norm and use a cutoff distance of $c = \unit[10]{m}$. In our case, the two sets $X,Y$ represent the ground truth target set and the estimated target set (from the PHD filter).

Since the OSPA can fluctuate due to false negative and false positive detections, we use the median of the OSPA over the last \unit[50]{s} of each trial to obtain a steady state estimate in each experiment.
From our previous work \cite{xin2022comparing, xin2024comparison}, we typically observe a rapid decline in the OSPA value at the outset, which then gradually levels off to a steady value as exploration progresses. Figure~\ref{fig:pureFittingresult} shows these trends for different $\theta$ values.

\subsubsection{Exploration Inefficiency}
We also introduce another metric, Exploration Inefficiency ($EI$), which evaluates the percentage of the unexplored area by the robot team over time:
\begin{equation}
EI = \left(1 - \frac{\text{Explored Area}}{\text{Total Area}}\right) \times 100\%
\label{eq:exploration_inefficiency}
\end{equation}
This metric will help in evaluating the efficiency of a MRS deployed for environmental monitoring tasks. A lower ($EI$) score suggests that a larger portion of the area has been explored, signifying higher efficiency in monitoring tasks. Therefore, in the context of ensuring comprehensive environmental surveillance, striving for a minimal Exploration Inefficiency percentage is key to maximizing the area monitored by the MRS.

\section{Problem Formulation}
This section is organized into three parts. Firstly, we introduce the fitting models used to predict the system's performance over time, where the model parameters are functions of the dimensionless variables. Secondly, we describe the method for generating dimensionless variables from the MRS parameters outlined in the previous section, utilizing the Buckingham Pi theorem. Thirdly, we explain the optimization method employed to determine the parameters for both the fitting model and the generation of dimensionless variables.

\subsection{Model Fitting}
\label{sec:model_fitting}
The objective of deploying fitting models is to succinctly describe the system's behavior over time using a reduced set of parameters. The general objective function is formulated as follows:
\begin{equation}
    \min\sum_{n=1}^{N}\sum_{t=1}^{T} \|\hat{m}(t \mid \theta_n) -m_{t,n} \|^2
    \label{eq:Objective Function}
\end{equation}
where $m_{t,n}$ is the value of the performance metric (OSPA or EI) at time $t$ in training data $n$ and $\hat{m}(t \mid \theta_n)$ denotes the prediction model for that metric, which depends on the team/task parameters $\theta_n$ in the training sequence $n$. 

The approach we use is to select a parametric model for $\hat{m}(t \mid \theta_n)$ and to try to connect the parameters of that model to $\theta_n$.
This will allow us to predict the performance based on the specific team parameter configurations at any given time. 
We introduce symbol $\Phi$ to denote the collection of model parameters in $\hat{m}(t \mid \theta_n)$.
Based on our observations in \cref{fig:pureFittingresult}, we will utilize two different models within the exponential family due to their inherent capability to encapsulate the decay present in the data.

The first is an exponential model:
\begin{equation}
\hat{m}(t \mid \Phi(\theta)) = a e^{-b t} + c
\label{eq:Exponential Function}
\end{equation}
Here the model parameter set is $\Phi = \{a, b, c\}$, which describe the initial state ($a$), decay rate ($b$), and steady-state behavior ($c$).
The second is a sigmoid model:
\begin{equation}
\hat{m}(t \mid \Phi(\theta)) = \frac{L}{1 + e^{-k t}} + d
\label{eq:sigmoid model}
\end{equation}
Here the model parameter set is $\Phi = \{L, k, d\}$, which describe the maximum value ($L$), the curve's steepness ($k$), and the baseline adjustment ($d$).

\begin{figure*}[tbph]
    \centering
    \subfloat[$\Phi = \{a, b, c\}$ vs $n_t$]{\includegraphics[width=0.45\linewidth]{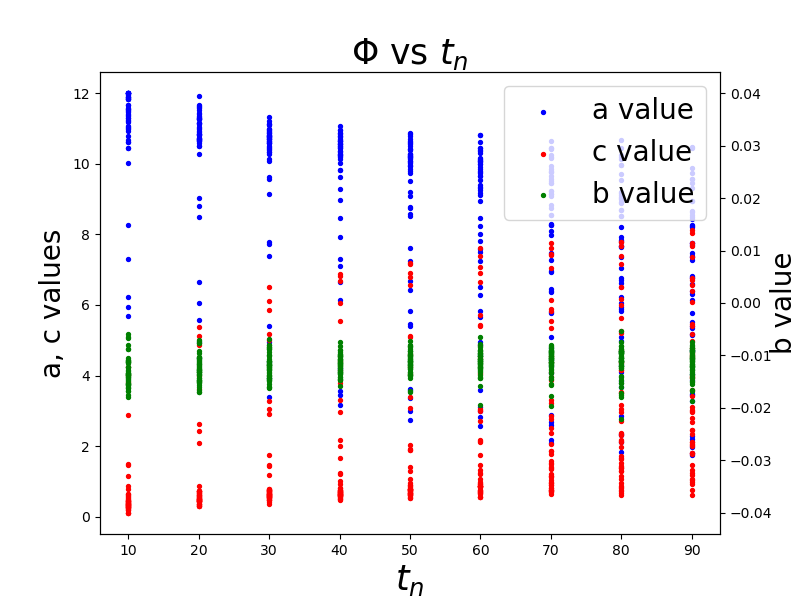}\label{fig:acoExponentialTargetvsabc}}
    \subfloat[$\Phi = \{L, k, d\}$ vs $n_t$]{\includegraphics[width=0.45\linewidth]{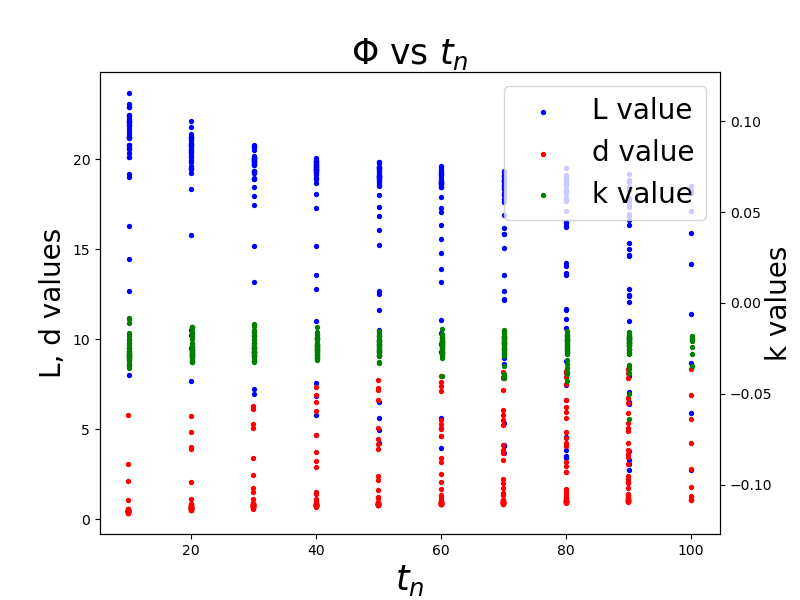}\label{fig:acoSigmoidTargetvsLKB}}
    \caption{Comparative analyses of the variables $a$, $b$, and $c$ from the Exponential fitting method, and $L$, $k$, and $d$ from the sigmoid fitting method, against the target number $n_t$. The data is from ACO method.}
    \label{fig:acoFittingComparisons}
\end{figure*}

These parameters play a critical role in modeling the predictive trajectory of system performance over time, capturing the system's dynamics effectively.
The most straightforward method to predict $\hat{m}(t \mid \Phi(\theta))$ involves first modeling $\Phi(\theta)$ with respect to $\theta$ and then utilizing $\Phi(\theta)$ for prediction. However, early tests indicate that directly modeling $\Phi(\theta)$ with $\theta$ poses challenges. Figures \ref{fig:acoExponentialTargetvsabc} and \ref{fig:acoSigmoidTargetvsLKB} display the values of the parameters $\Phi$ against the number of targets $n_t$, using the exponential and sigmoid fitting methods, respectively. While there is a discernible trend, a large variance in the data makes accurate prediction difficult since you can have vastly different $\Phi$ values for the same $n_t$. Similar trends hold for the other parameters within $\theta$.


\subsection{Dimensionless Variable Structure}
\label{sec:dimensionless_variables}
In this section, we propose the formula of a dimensionless variable $\Pi(\theta)$ for the MR-MTT problem. The idea to depict system performance through a dimensionless variable is influenced by preceding research, such as Xie et al.~\cite{Xie2022}, which utilizes dimensionless variables to model physical laws. Moreover, the study by Kwa et al.~\cite{kwa2023effect} also lends inspiration, investigating the correlation between robot density and the phase transition from exploration to exploitation within MRS.
The primary benefit of a dimensionless variable $\Pi(\theta)$ is to enable comparisons across different scales. 

To find potential dimensionless variables, we first create a dimensional matrix $\mathbf{D}$. 
In such a matrix, there is a column for each a variable and a row for each type of unit.
In our case, we have $n = 5$ dimensional variables (from $\Pi = (n_r, n_t, r, \rho_r, \rho_t)$) with $m = 2$ units (item and meter).
The entries in row $i$ and column $j$ is the power of unit $i$ in variable $j$, as shown below.
For example, we can read column 4 as saying that the variable $\rho_r$ has units of quantity per square meter.
\begin{equation}
    \begin{gathered}
    \bf{D} = 
        \begin{blockarray}{cccccc}
            \text{$n_r$} & \text{$n_t$} & \text{$r$} & \text{$\rho_r$}  & 
             \text{$\rho_t$} & \\
            \begin{block}{[ccccc]c}
                1 & 1 & 0 & 1 & 1 & \text{unit} \\
                0 & 0 & 1 & -2 & -2 & \text{meter}\\
            \end{block}
        \end{blockarray}
    \end{gathered}
\end{equation}

A dimensionless variable is one such that the multiplication (and division) of its constituent parts has no resultant units. In our case, we seek a combination of the columns of $\mathbf{D}$ such that the resultant is the zero vector (meaning it has no units of any kind).
From linear algebra, this is the null space of $\mathbf{D}$.
According to the Buckingham Pi theorem \cite{curtis1982dimensional}, there are $n - m = 3$ vectors necessary to span the null space of $\mathbf{D}$, represented by $\mathbf{W} = [\mathbf{w_1}, \mathbf{w_2}, \mathbf{w_3}]$, as follows:
\begin{equation}
    \bf{W} = [\bf{w_{1}}, \bf{w_{2}}, \bf{w_{3}}]=
    \begin{bmatrix} 
        -1 & -1 & -1\\
        1 & 0 & 0\\
        0 & 2 & 2\\
        0 & 1 & 0\\
        0 & 0 & 1\\
    \end{bmatrix}
\end{equation}
These vectors form a basis for the null space of $\mathbf{D}$.
Finally, we can assert that any linear combination $\bf{w}$ of these basis vectors will also be dimensionless:
\begin{equation}
\begin{split}
    \bf{w} &= \gamma_1 \bf{w_{1}} + \gamma_2 \bf{w_{2}} + \gamma_3 \bf{w_{3}}\\
          &= [w_{n_r}, w_{n_t}, w_r, w_{\rho_r}, w_{\rho_t}]^T 
\end{split}
\label{eq:w=[w_b1,w_b2,w_b3]}
\end{equation}

For our variable set $\theta$, any dimensionless variable $\Pi(\theta)$ can thus be written in the following forms:
\begin{subequations}\label{eq:dimensionless_variable_structure}
\begin{align}
   \Pi(\theta, \mathbf{w}) 
   &= n_r^{w_{n_r}}n_t^{w_{n_t}}r^{w_r}\rho_r^{w_{\rho_r}}\rho_t^{w_{\rho_t}} \label{eq:omega_form}\\
   \Pi(\theta, \boldsymbol{\gamma}) &= \left(\frac{n_t}{n_r}\right)^{\gamma_1} \left(\frac{r^2 \rho_r}{n_r}\right)^{\gamma_2} 
   \left(\frac{r^2 \rho_t}{n_r}\right)^{\gamma_3} \label{eq:gamma_form}
\end{align}
\end{subequations}
We will call \eqref{eq:omega_form} the $\mathbf{w}$ form of $\Pi$ and call \eqref{eq:gamma_form} the $\gamma$ form of $\Pi$.
The goal is to learn the coefficients $\boldsymbol{\gamma} = [\gamma_1, \gamma_2, \gamma_3]^T$ to find a ``useful'' $\Pi$ that uncovers structure in the MRS performance. 

A crucial aspect of our analysis is the examination of the structure of the dimensionless variables within these models. Specifically, we focus on the powers assigned to each component parameter ($\bf{w}$ or $\boldsymbol{\gamma}$), which represent the ``structure'' of the dimensionless variable $\Pi(\theta)$. If we find a similar structure regardless of the fitting model used, it will lend greater credibility to the idea that the dimensionless variable provides meaningful insights into the underlying performance of the MRS system.

\subsection{Procedure for Learning Dimensionless Variable}
\label{sec:comparison_analysis}
With the general structure of the dimensionless variable $\Pi(\theta)$ at our disposal, we now seek to co-optimize the structure of $\Pi$ and the relationship between $\Pi$ and the coefficients $\Phi$ of our performance models. In other words, for the exponential model we want to find \{$a(\Pi(\theta))$, $b(\Pi(\theta))$, and $c(\Pi(\theta))$\}. In fitting parameters within both the exponential and sigmoid models, we utilize the \textit{same} dimensionless variable $\Pi(\theta)$. This approach ensures consistency across models when optimizing the different parameters for model $\hat{m}$. 

Based on the literature \cite{Xie2022}, we employ a polynomial fitting approach:
\begin{equation}
    \phi(\Pi(\theta), \boldsymbol{\beta}_\phi) = \Sigma_{k=0}^{K} \beta_{\phi,k}\Pi(\theta)^{k},
\label{eq:polynomial fitting}
\end{equation}
where $\phi \in \Phi$ is any of the model parameters, \eg $a, b, c$ for the exponential model, and $\beta_{\phi,k} \in \boldsymbol{\beta}_\phi$ are the coefficients for the polynomial representing $\phi$.
We denote $\boldsymbol{\beta} = \{ \boldsymbol{\beta}_\phi \mid \phi \in \Phi \}$ as the collection of all the $\beta$ parameters.
It is important to consider that higher-order polynomial functions may lead to overfitting. In our case, we limit our polynomial functions to fourth-order (\ie $K=4)$. 

By integrating equations \eqref{eq:dimensionless_variable_structure} and \eqref{eq:polynomial fitting}, we can efficiently optimize the objective function outlined in equation \eqref{eq:Objective Function}, which we can rewrite as:
\begin{equation}
\argmin_{\boldsymbol{\beta},\boldsymbol{\gamma}} \sum_{n=1}^{N} \sum_{t=1}^{T} \left\| \hat{m}\Big(t \mid \Phi\big(\Pi(\theta_n, \boldsymbol{\gamma}), \boldsymbol{\beta}\big)\Big) - m_{t,n} \right\|^2.
\label{eq:objective_final}
\end{equation}
The decision variables are the $\boldsymbol{\beta}$ in equation \eqref{eq:polynomial fitting} and the $\boldsymbol{\gamma}$ in equation \eqref{eq:dimensionless_variable_structure}.
This enables the model to capture non-linear relationships between the input MRS parameters and the system performance.

\begin{algorithm}[tbph]
    \caption{Dimensionless Variable Analysis}
    \begin{algorithmic}[1]
        \State Initialize \textbf{D} and find its null space \textbf{W}
        \State Initialize $\boldsymbol{\gamma}$ and $\boldsymbol{\beta}$
        \While{MSE from $\eqref{eq:objective_final} > 10^{-4}$}
            \State Calculate $\Pi(\theta, \boldsymbol{\gamma})$ using \eqref{eq:gamma_form}.
            \State Calculate $\Phi(\Pi, \boldsymbol{\beta})$ using \eqref{eq:polynomial fitting}.
            \State Predict system performance using \eqref{eq:Exponential Function} or \eqref{eq:sigmoid model}.
            \State Apply gradient descent on \eqref{eq:objective_final} to update $\boldsymbol{\gamma}$.
            \State Predict system performance using equation \eqref{eq:Exponential Function} or \eqref{eq:sigmoid model}.
            \State Apply gradient descent on \eqref{eq:objective_final} to update $\boldsymbol{\beta}$.
        \EndWhile
        \State return $\boldsymbol{\gamma}$, $\boldsymbol{\beta}$
    \end{algorithmic}
    \label{algo:Dimensionless Variable Analysis pseudo code}
\end{algorithm}

Algorithm \ref{algo:Dimensionless Variable Analysis pseudo code} outlines our approach to minimize the objective function \eqref{eq:objective_final}.
It starts by initializing a dimensional matrix $\textbf{D}$ and calculating its null space $\textbf{W}$. Initial guesses for the parameters $\boldsymbol{\gamma}$ and $\boldsymbol{\beta}$ are established; a well-chosen initial guess can significantly reduce convergence time. 
Before running the training process, we first use a set of random parameters for pre-training and obtain a new set of parameters as good initial guess parameters for formal training.

The algorithm then iteratively refines these parameters until convergence is achieved.
Each iteration involves calculating the dimensionless variable $\Pi(\theta)$, followed by the determination of fitting model parameters $\boldsymbol{\beta}$ based on polynomial fitting. The system performance is then predicted using either the exponential function \eqref{eq:Exponential Function} or the sigmoid model \eqref{eq:sigmoid model}. Gradient descent is applied to update the parameters $\boldsymbol{\gamma}$ and $\boldsymbol{\beta}$ by minimizing the mean squared error between the predicted and actual system performances, as defined by the objective function. 

We use Adaptive Moment Estimation (Adam) \cite{KingBa15} within PyTorch. Adam optimizer is a powerful and general optimization approach used for high-dimensional optimization in machine learning. The two-step optimization method is used because it allows for a more flexible and powerful model than a linear model. By optimizing the $\boldsymbol{\beta}$ and $\boldsymbol{\gamma}$ parameters separately, each step can focus on a different aspect of the model, making the optimization process more efficient.
This is similar to the expectation maximization (EM) procedure commonly used in machine learning.

\section{Results Analysis}
To demonstrate the utility of our approach, we learn the $\boldsymbol{\beta}$ and $\boldsymbol{\gamma}$ using the dataset from \cref{sec:dataset}. Note that for each search algorithm (Lloyd's, SA, PSO, ACO, and AIS), each metric (OSPA and EI), and each model type (exponential and logistic), we learn a separate $\Pi$, resulting in 20 dimensionless variable results. 
Note that within a specific search algorithm, metric, and model, all $\Phi$ for that model use the same $\Pi$.
For each of the 20 datasets, we split the dataset into two disjoint pieces: a training set (350 samples) and a test set (150 samples). Only the training set is used to optimize \eqref{eq:Objective Function} using \cref{algo:Dimensionless Variable Analysis pseudo code} and the test set is used to ensure the generalizability of the results.

\begin{table*}[tbph]
  \centering
  \caption{Comparison of MSE Using the OSPA Metric}
  \label{tab:comparison_ospa}
  \begin{tabular}{l|cc|cccc}
    \toprule
     & \multicolumn{2}{c|}{Model Fitting Error}  & \multicolumn{4}{c}{Dimensionless Variable Prediction Error}    \\ \cline{2-7}
                     & Exponential & Sigmoid   
                     & \multicolumn{2}{c}{Exponential} & \multicolumn{2}{c}{Sigmoid}    \\ \cline{4-7}
    MR-MTT Algorithm                 & &          
                     & Train          & Test           & Train        & Test            \\

    \midrule
    Lloyd's & 0.050  & 0.035   & 0.3631  &0.3770  & 0.4691  & 0.4761   \\
    PSO     & 0.079  & 0.073   & 0.2442  &0.2586  & 0.2982  & 0.3360  \\
    ACO     & 0.134  & 0.110   & 0.2914  &0.2932  & 0.3589  & 0.3663  \\
    AIS     & 0.095  & 0.079   & 0.2854  &0.3108  & 0.3458  & 0.4227  \\
    SA      & 0.058  & 0.072   & 0.1893  &0.2126  & 0.2563  & 0.3296  \\
    \bottomrule
  \end{tabular}
\end{table*}

\subsection{Results with OSPA Metric}
We first measure the ability of our models $\hat{m}(\Phi)$ to fit trends in the OSPA data from \eqref{eq:OSPA} over time.
We do this in two ways: 1) finding the best fit curve for each sample in the dataset (\ie each MR-MTT trial) and 2) using the dimensionless variable to fit $\Phi(\Pi(\theta))$.
This allows us to see how well our learned model is able to generalize over the full dataset.

\subsubsection{Single Sample Model Fitting}
We first measure the accuracy of fitting the exponential and sigmoid models $\hat{m}(\Phi_n)$ directly to the data. In this case, we fit a unique $\Phi_n$ for each sample $n \in 1, \ldots N$ in the dataset (\ie each of the $N=500$ MR-MTT trials). This baseline is to demonstrate that both models can accurately fit the data.
Results for this are shown in the Model Fitting Error columns in \cref{tab:comparison_ospa}.
We see that the MSE values are small relative to the maximum value of $c=10$ for OSPA, demonstrating a good ability to capture the trend in the data.

We also see that the sigmoid model \eqref{eq:sigmoid model} is slightly better at fitting the OSPA data than the exponential model \eqref{eq:Exponential Function}, except in the case of SA. This implies that the sigmoid method has a better fit for the OSPA metric in these MR-MTT algorithms. This could be due to the sigmoid function's ability to better capture the saturation effect seen in tracking performance over time.

\begin{figure*}[tbph]
\centering
\includegraphics[width=1\textwidth]{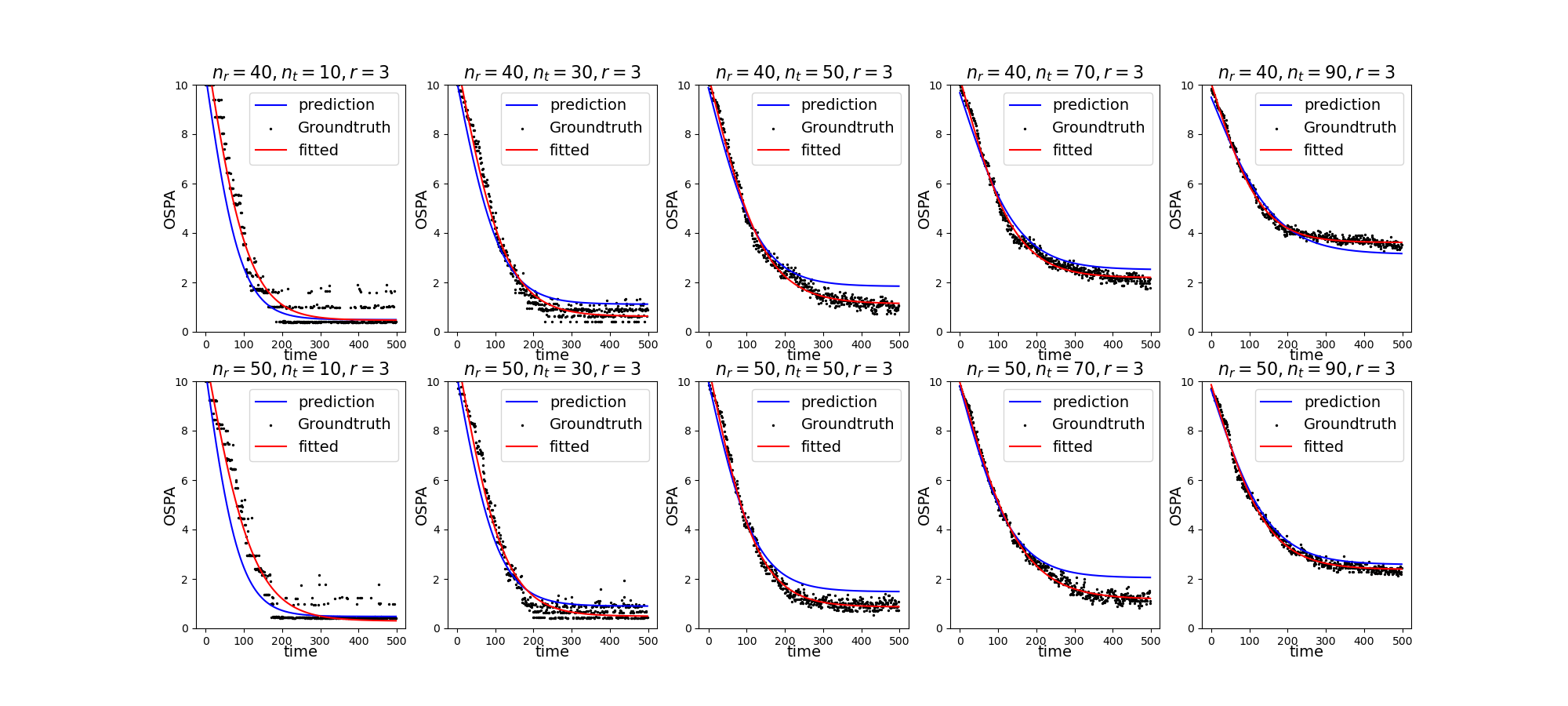}
\caption{Comparison between groundtruth OSPA, sigmoid fitted OSPA and predicted OSPA via dimensionless variable from PSO method.}
\label{fig:psoComparisonSigmoid}
\end{figure*}

\begin{figure*}[tbph]
\centering
\includegraphics[width=1\textwidth]{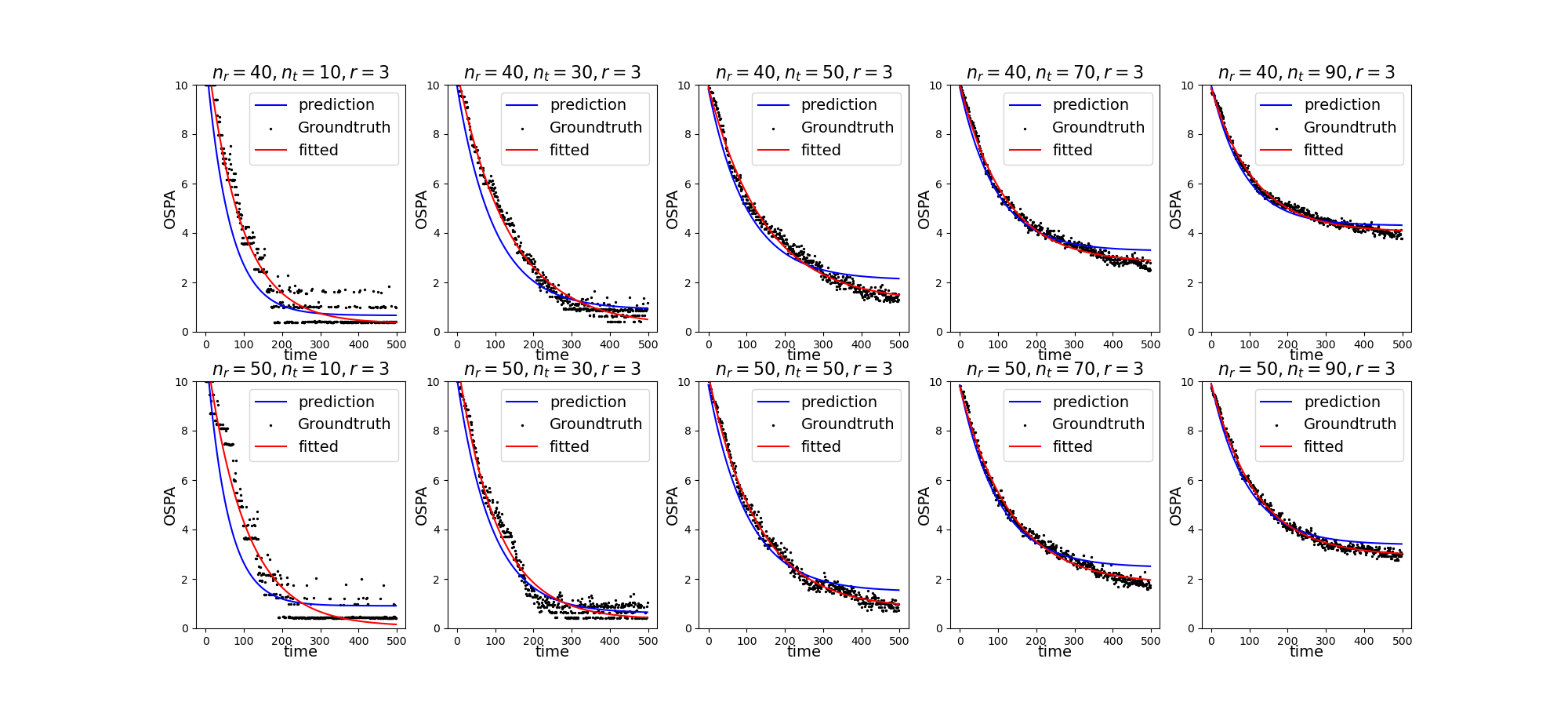}
\caption{Comparison between groundtruth OSPA, exponential fitted OSPA and predicted OSPA via dimensionless variable from SA method.}
\label{fig:saComparisonExponential}
\end{figure*}
 
We can also see the actual fit models against the data. 
The red lines in \cref{fig:psoComparisonSigmoid} shows results for the PSO search algorithm and sigmoid model and \cref{fig:saComparisonExponential} shows results for the SA search algorithm and the exponential model.
We can again see that the models accurately represent each sample in the dataset.
In both cases, we see that the OSPA decreases more slowly as the number of targets $n_t$ increases and that the OSPA reaches a lower value at the same time when the number of robots $n_r$ increases.

\subsubsection{Dimensionless Variable Model Fitting}
Next, we utilize the dimensionless variable to fit $\hat{m}(\Phi(\Pi(\theta)))$, where the parameters $\phi \in \Phi$ depend upon the dimensionless variable $\Pi(\theta)$ which depends upon the team/task parameters in $\theta$.
Results are shown in the Dimensionless Variable Prediction Error columns of \cref{tab:comparison_ospa} and the blue lines in \cref{fig:psoComparisonSigmoid,fig:saComparisonExponential}.
We see that the results are slightly worse than directly fitting $\Phi_n$ to each individual sample. This is to be expected, as the previous results essentially ``overfit'' to each sample and offer no ability to generalize to new scenarios $\theta$.

\begin{figure}[htbp]
    \centering
    \includegraphics[width=0.5\textwidth]{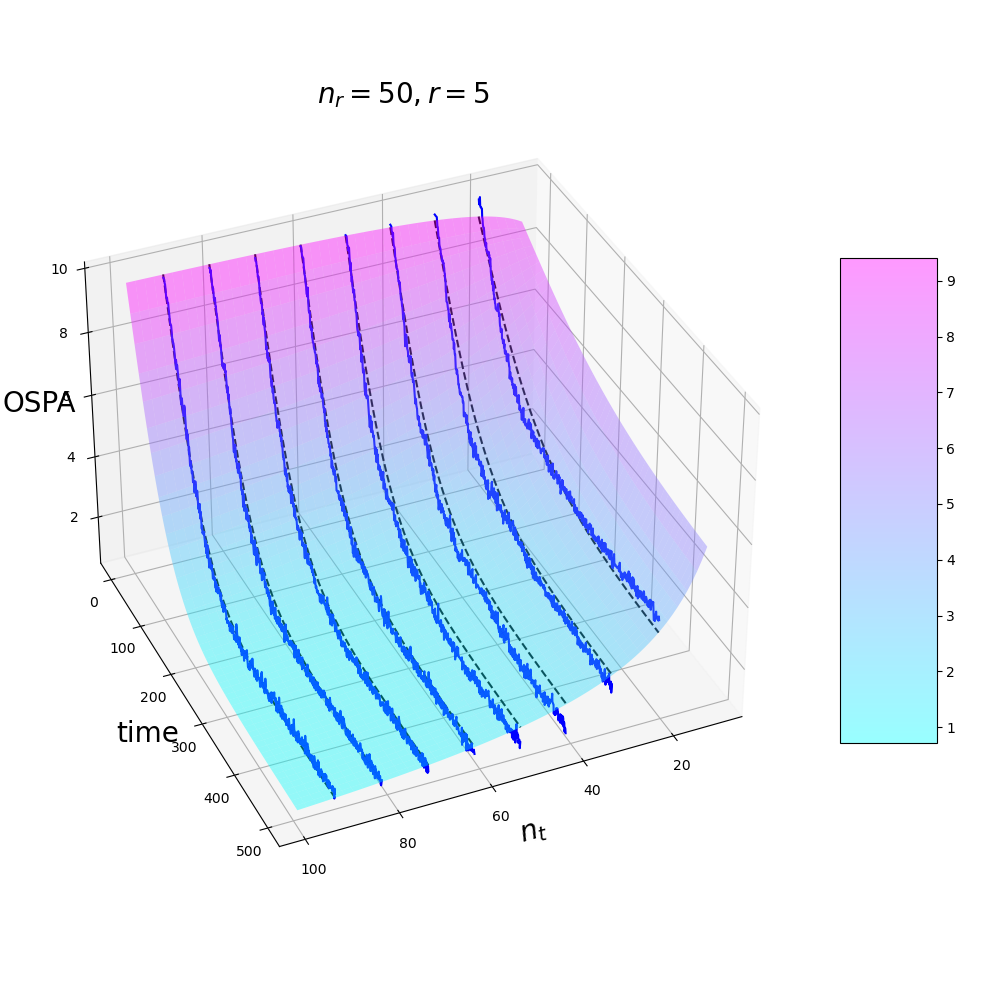}
    \caption{The surface plot illustrates the evolution of OSPA values for a fixed $n_r=50$ and $r=5$. The blue solid lines on the surface denote the actual observed OSPA values (`ground truth') for target numbers ranging from 10 to 100 and the black lines are the predicted model on the surface.}
    \label{fig:SAr50r5-3d}
\end{figure}

We can see that the predicted (blue) lines in \cref{fig:psoComparisonSigmoid,fig:saComparisonExponential} are slightly less accurate that the direct fit (red) lines, but the blue lines still capture the essential trends in the data.
The MSE in \cref{tab:comparison_ospa} shows that our predictions are still reasonably accurate, with errors less than 5\% of the maximum OSPA value.
More importantly, we see that the MSE on the test data is quite similar to that on the training data, demonstrating the ability of our method to accurately interpolate and extrapolate to scenarios outside of the training data.
These results hold for all the different search algorithms and both fit models, demonstrating the generalizability of the prediction methodology.

\Cref{fig:SAr50r5-3d} shows an example of the interpolation between scenarios $\theta$ using our model. Here, we visualize the predicted OSPA as a function of time across a range of target numbers $n_t$ (with fixed robot count $n_r$ and sensing radius $r$). We see that the fit model is generally quite close to the training data (blue lines).
We also note the gradient of performance visible along the $n_t$ axis: when there are fewer targets, the system quickly converges to a low OSPA score, which corresponds to better tracking performance. In scenarios with many targets, the OSPA score decreases more slowly and reaches the steady state early.
We also see that our model yields a smooth transition in system performance degradation as $n_t$ increases. This continuity is expected since the rate of change in system performance should not experience abrupt shifts with incremental variations in $n_t$.
The accuracy of the predictions and the smooth performance gradient both show the utility of the dimensionless variable approach in system performance prediction.

Using the dimensionless variable, the prediction errors associated with the exponential model are generally lower than those of the sigmoid model for both training and testing datasets across all search algorithms. Despite the sigmoid model being a better fit for the single-sample case, the exponential model may be more advantageous for predictions via dimensionless variable.

\begin{figure*}[tbph]
    \centering
    \subfloat[$\Phi = \{a, b, c\}$ vs $\Pi(\theta)$]{\includegraphics[width=0.5\linewidth]{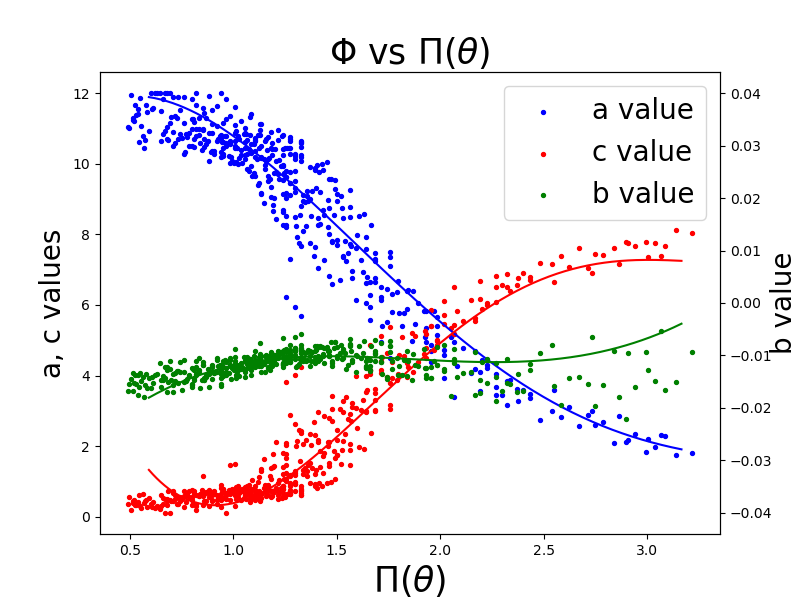}\label{fig:acoExponentialDMvsabc}}
      \hfill 
  \subfloat[$\Phi = \{L, k, d\}$ vs $\Pi(\theta)$]{\includegraphics[width=0.5\linewidth]{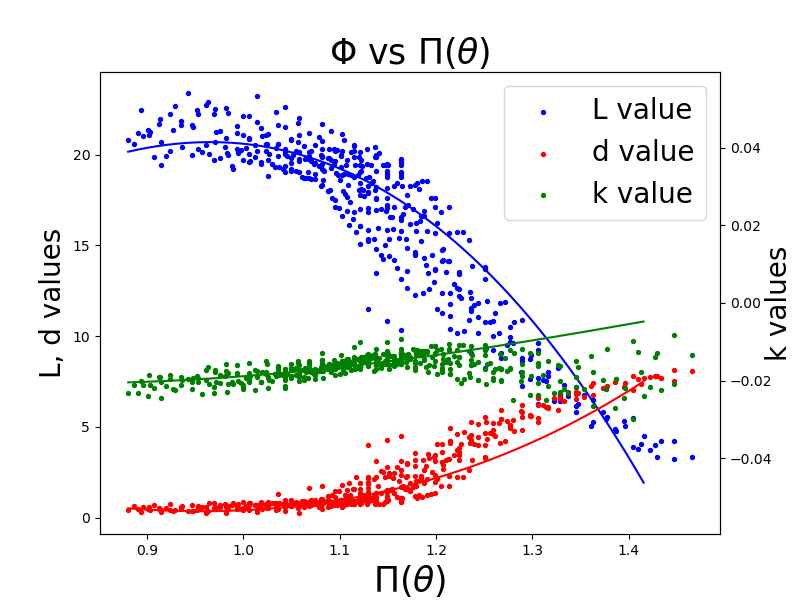}\label{fig:acoSigmoidDMvsLKB}}

  \caption{Comparative analysis of the variables L, k, and d from the Sigmoid fitting method, and a, b, and c from the Exponential fitting method, against the dimensionless variable $\Pi(\theta)$. The data is collected from ACO method.}
  \label{fig:acoComparisons}
\end{figure*}

\subsubsection{Model Parameter Analysis}
As we saw in \cref{fig:acoFittingComparisons}, individual parameters in $\theta$ show hints of trends in performance but there is too much noise to allow the method to capture the true trends.
\Cref{fig:acoComparisons} shows the same results using our dimensionless variable analysis of the system.
In the figures, the dots are the samples (\ie the single-sample fits from the dataset) and the lines are the best fit polynomials for $\phi(\theta, \boldsymbol{\beta}_\phi)$ from \eqref{eq:polynomial fitting}.
Here, we see that the learned dimensionless variable $\Pi$ is able to find clear structure in how the parameters of the model $\Phi$ vary based on $\Pi$.
The findings suggest that the dimensionless variable $\Pi(\theta)$ has a strong predictive capability for the parameters across both fitting methods.

Investigating the trends, we see from both plots that the rate of decay ($b$ and $k$) has small variation of the range of $\Pi$. However, the steady-state value ($c$ and $d$) both increase with $\Pi$ while the scale of the error ($a$ and $L$) have a simultaneous decrease. This makes sense, as OSPA always starts at 10 so $a+c$ or $\frac{L}{2}+d$ should always be $\approx 10$.


\begin{table*}[tbph]
  \centering
  \caption{Comparison of MSE Using the EI Metric}
  \label{tab:comparison_ei}
  \begin{tabular}{l|l|cc}
    \toprule
     &  & \multicolumn{2}{c}{Dimensionless Variable Prediction} \\ \cline{3-4}
    MR-MTT Algorithm & Model Fitting Error & \multicolumn{1}{c}{Train} & \multicolumn{1}{c}{Test} \\
    \midrule
    Lloyd's & 0.00032   & 0.00478 & 0.00507\\
    ACO & 0.00062 & 0.00244 & 0.00282\\
    AIS & 0.00043 & 0.00202 & 0.00220\\
    PSO & 0.00028 & 0.00175 & 0.00295\\
    SA  & 0.00034 & 0.00185 & 0.00289\\
    \bottomrule
  \end{tabular}
\end{table*}

\subsection{Results with EI Metric}
We repeat the above analysis for the EI metric from \eqref{eq:exploration_inefficiency}, where we only show results for the sigmoid model as it has much lower error than the exponential model.
Both metrics typically exhibit a rapid decrease at the start, then the rate of improvement in both metrics tends to slow over time. However, OSPA is more sensitive to the dynamics and behaviors of the targets being tracked, as well as the noise in the robots' sensing. On the other hand, EI is more influenced by the distribution of robots in the environment and their mobility, as it focuses on the spatial aspect of the task.

\begin{figure*}[t]
\centering
\includegraphics[width=1\textwidth]{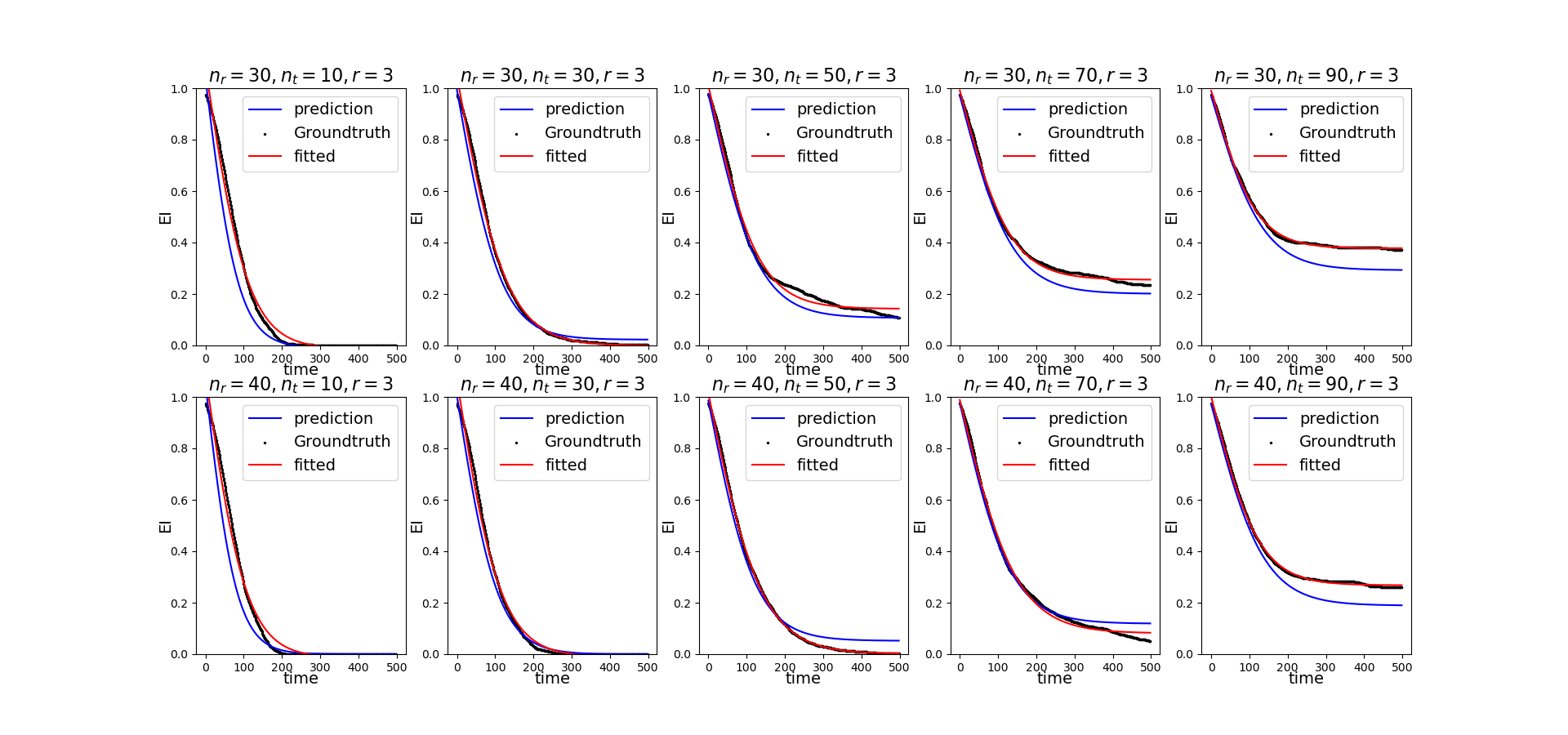}
\caption{Comparison between groundtruth EI, sigmoid fitted EI and predicted EI via dimensionless variable from PSO method.}
\label{fig:pso_undetectedAreaSigmoid}
\end{figure*}

\Cref{tab:comparison_ei} show the MSE for the single-sample fitting and the dimensionless variable fitting.
Again, we see that the sigmoid model fits the data well, with errors less than 0.5\% of the maximum value of 1 for EI in all cases.
This decrease in relative error is due to the stability of EI versus OSPA as it does not depend on sensor noise.
We can visualize these fits in \cref{fig:pso_undetectedAreaSigmoid}.

\begin{figure}[t]
\centering
\includegraphics[width=0.5\textwidth]{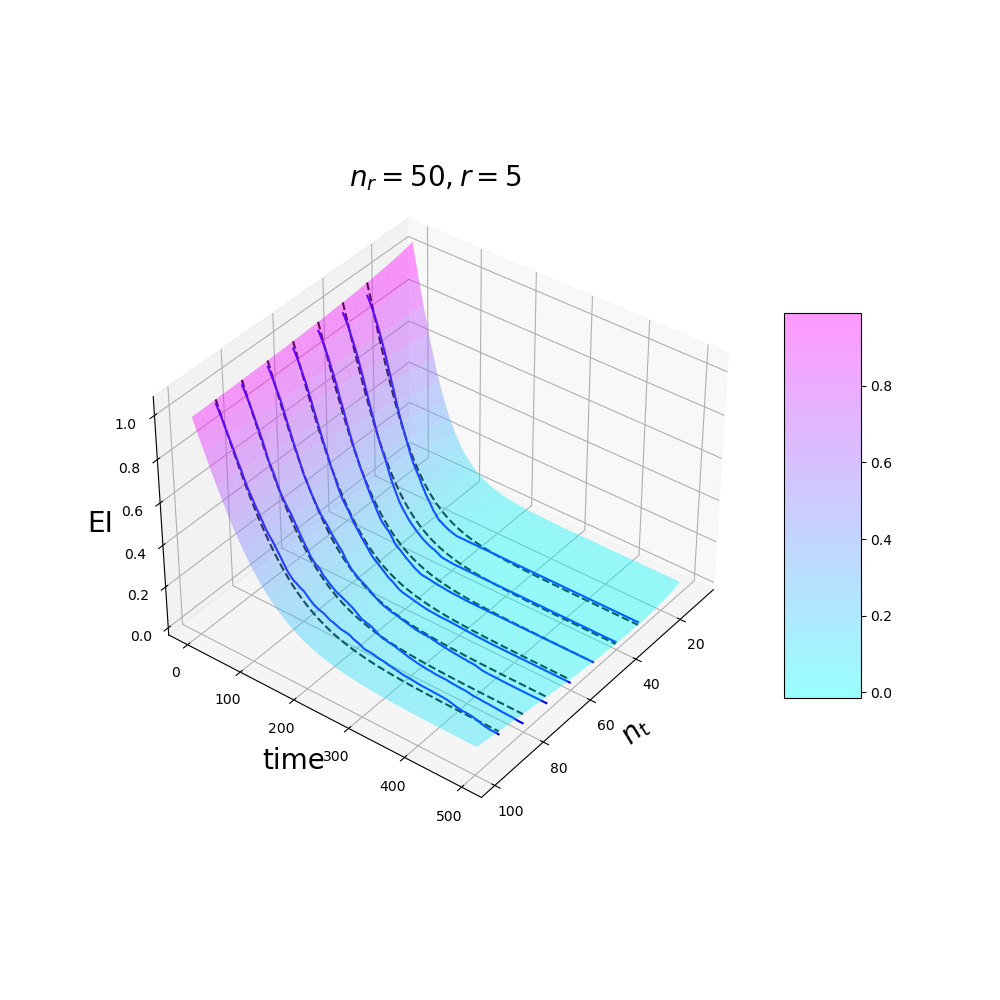}
\caption{The surface plot illustrates the evolution of EI values for a fixed fixed $n_r=50$ and $r=5$. The blue solid lines on the surface denote the actual observed EI values (`ground truth') for target numbers ranging from 10 to 100 and the black lines are the predicted model on the surface.}
\label{fig:psor50r5-3d-UndetectedArea}
\end{figure}

We again see that the dimensionless variable errors are higher than the single-sample fits, as is expected, and that the MSE on the test data is only slightly higher than on the training data, demonstrating the ability of our method to accurately interpolate and extrapolate to new scenarios outside of the training set.
\Cref{fig:psor50r5-3d-UndetectedArea} shows a typical example of this interpolation.

\subsection{Dimensionless Variable Structure}
The results above all demonstrate the utility of our method to accurately fit data and to interpolate or extrapolate to new scenarios $\theta$ outside of the training set.
Our methodology has an additional benefit in terms of MRS system-level analysis in that the structure of the dimensionless variable $\Pi$ can provide insight into the structure of the task and identify the most important factors that affect system performance.
Recall that we had two forms of the dimensionless variable, the $\mathbf{w}$ form from \eqref{eq:omega_form} and the $\boldsymbol{\gamma}$ form from \eqref{eq:gamma_form}.

\subsubsection{$\mathbf{w}$ Form}
The $\mathbf{w}$ form of $\Pi$ allows us to examine the effects of each individual parameter in $\theta$ on the predicted team performance. 
Recall the definition of $\Pi(\theta, \mathbf{w}) = n_r^{w_{n_r}}n_t^{w_{n_t}}r^{w_r}\rho_r^{w_{\rho_r}}\rho_t^{w_{\rho_t}}$.
Here, the values in $\mathbf{w}$ are the exponents for each variable in $\theta$. This means that a larger absolute value $|w|$ indicates that changes in that parameter will have a larger impact on the resulting value of $\Pi$ (where $|w|=0$ means that the variable has no effect on the resulting performance).

\begin{figure*}[tbph]
    \centering
    \subfloat[OSPA Metric]{\includegraphics[width=0.45\linewidth]{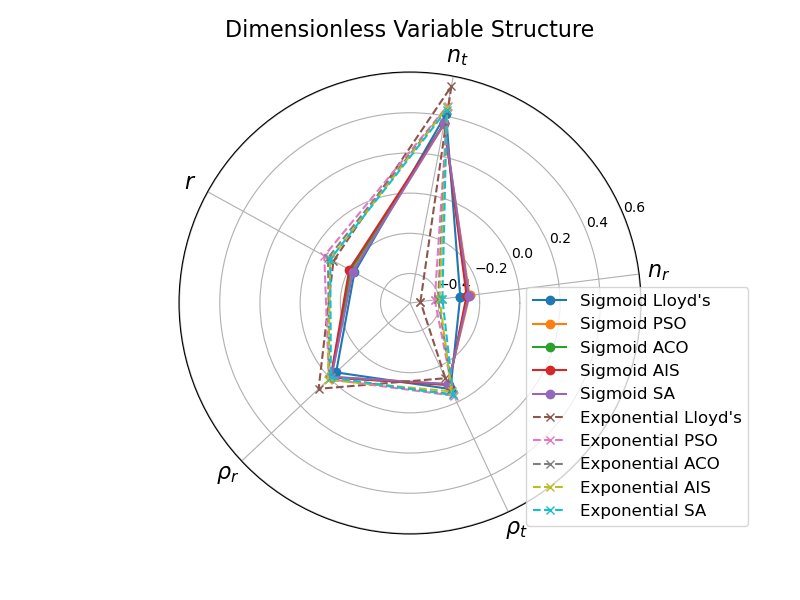}\label{fig:DMSigmoidExponentialOSPA}}
    \subfloat[EI Metric]{\includegraphics[width=0.45\linewidth]{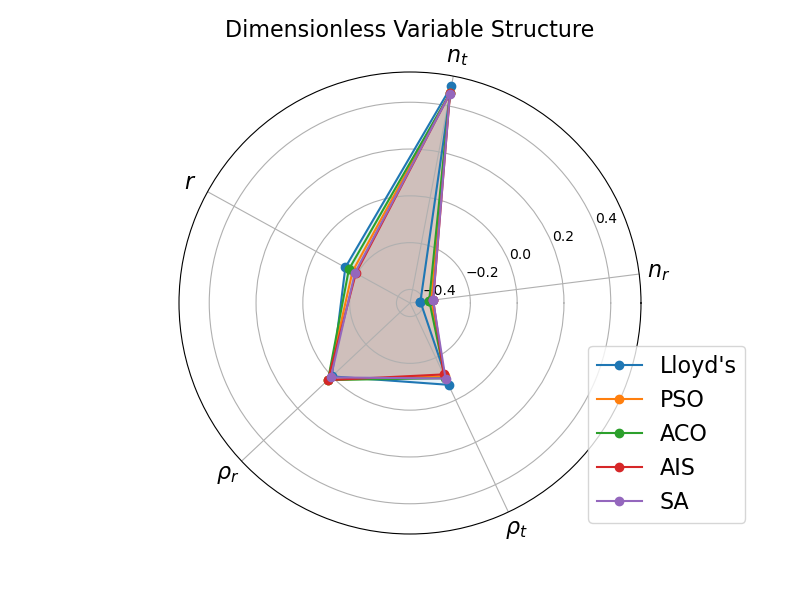}\label{fig:DMUndetectedArea}}
    \caption{Dimensionless variable structure with sigmoid and exponential model fitting method across different MR-MTT algorithms.}
    \label{fig:DM structure}
    

\end{figure*}

To illustrate the structure of the dimensionless variable, we use a polar plot.
\Cref{fig:DMSigmoidExponentialOSPA} shows the structure of $\Pi$ for our OSPA data and \cref{fig:DMUndetectedArea} for the EI data.
We see several important trends.
First, we see that the number of targets $n_t$ and the number of robots $n_r$ have the largest effects on term performance, with $n_t$ having a large positive exponent ($\approx0.45$) and $n_r$ having a large negative exponent ($\approx0.35$). This means that the number of targets has the largest relative effect on team tracking performance.
Additionally, it means that increasing $n_t$ affects performance in the opposite direction that increase $n_r$ does, a result that agrees with intuition as having a larger team or a smaller target set should both affect tracking in the same way.
The sensing radius has the next most important effect, with exponents $\approx-0.2$ when using the sigmoid model and $\approx -0.1$ when using the exponential model.

In addition to providing insight into the relative importance of each term, we also see that the structure of the dimensionless variables is quite consistent across search algorithm, performance metric, and model family.
This indicates that our dimensionless variable truly is capturing an important essence of the MR-MTT task.

\begin{figure*}[tbph]
    \centering
    \subfloat[OSPA Metric]{\includegraphics[width=0.45\linewidth]{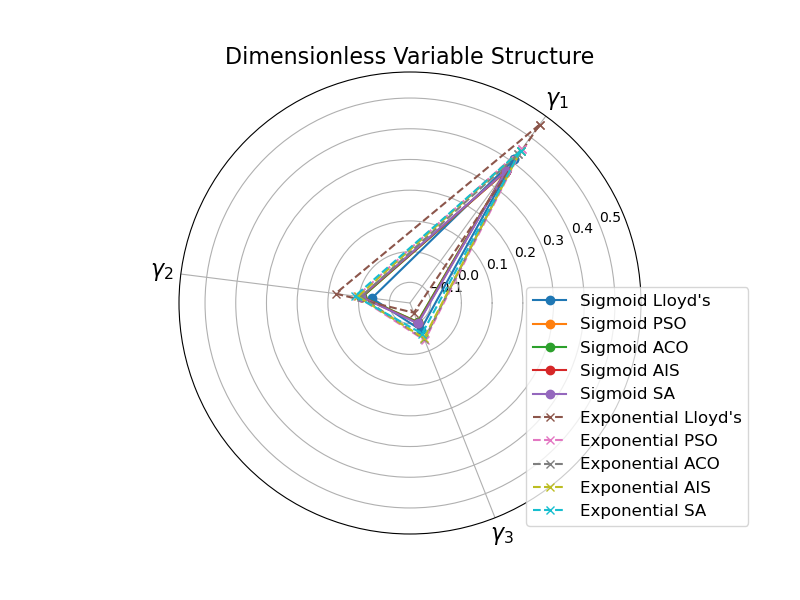}\label{fig:DMSigmoidExponentialOSPAGamma}}
    \subfloat[EI Metric]{\includegraphics[width=0.45\linewidth]{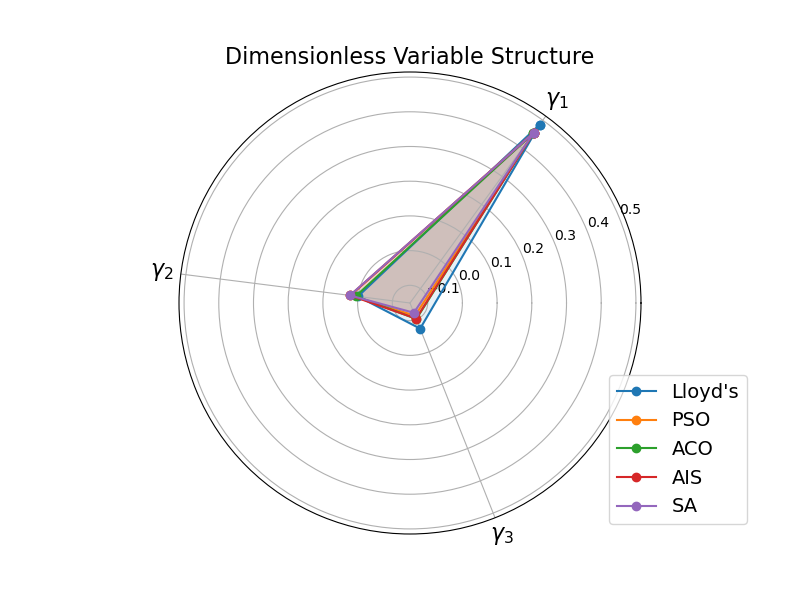}\label{fig:DMUndetectedAreaGamma}}
    \caption{Dimensionless variable structure represented by $\gamma$ value with sigmoid model fitting method across different MR-MTT algorithms.}
    \label{fig:DM structure Gamma}
\end{figure*}

\subsubsection{$\boldsymbol{\gamma}$ Form}
We can also examine the $\boldsymbol{\gamma}$ form of the dimensionless variable, $\Pi(\theta, \boldsymbol{\gamma}) = \left(\frac{n_t}{n_r}\right)^{\gamma_1} \left(\frac{r^2 \rho_r}{n_r}\right)^{\gamma_2}    \left(\frac{r^2 \rho_t}{n_r}\right)^{\gamma_3}$.
This analysis can show which of the bases for the null space $W$ of the dimension matrix $\mathbf{D}$ are most important.
\Cref{fig:DMSigmoidExponentialOSPAGamma,fig:DMUndetectedAreaGamma} shows the resulting polar plots for the values of each element in $\boldsymbol{\gamma}$.
We can see that $\gamma_1$ has the highest exponent by far, indicating that the ratio of robots to targets plays the most important role in determining the performance of the team.
The other two exponents are quite small, indicating that the ratio of sensing radius to environment size plays a much smaller role in describing the trends in the data.

\subsubsection{Discussion}
We see that dimensionless variables meaningfully encapsulate system properties, implying that for a given system, different dimensionless variables exhibit a similar structure due to the consistent influence of system parameters. 
This uniform structure of $\Pi(\theta)$ across different MR-MTT algorithms and parametric fitting models suggests that the impact of these dimensionless variables on system performance is consistent regardless of the specific MR-MTT algorithms and fitting model used. 
Additionally, small differences in exponents can imply which algorithms are more or less sensitive to changes in the team or task.

\section{Conclusion}
This work has presented an in-depth analysis of MRSs in MR-MTT tasks using dimensionless variables to simplify the analysis of the high-dimensional parameter space and enable meaningful comparisons across various system configurations. Dimensionless variables help to identify fundamental characteristics that influence system performance, such as the ratio of robot numbers to target numbers and their densities within the operational environment. Furthermore, the ability to predict performance across different MR-MTT algorithms facilitates the selection of the most effective algorithm when the value of system parameters is known.

The findings demonstrate that the use of dimensionless variables can significantly contribute to the understanding, comparison, and optimization of MR-MTT algorithms. Our analysis indicates that certain structures of these dimensionless variables are consistent across different algorithms and fitting methods, suggesting underlying principles that govern the dynamics of MR-MTT systems. This consistency reinforces the potential of dimensionless variables as a powerful tool for predicting system performance and aiding in the design and deployment of distributed MRS.
Moreover, our comparative analysis of fitting and prediction errors using both the OSPA and EI metrics reveals insights into the predictive capabilities over different evaluation metric with dimensionless variable.

Despite the successes demonstrated, the task of predicting MR-MTT system performance remains complex, due to the inherent uncertainties and the dynamic nature of real-world environments. While the current study has focused on MR-MTT systems, the approach and findings have broader implications for the field of distributed MRS and predictive systems in robots more broadly. So long as there is a system performance metric that can be described by a parameter model $\hat{m}$, our method could be used to predict performance. This highlights the potential benefits of employing dimensionless analysis in complex systems and underscore the need for continued research into robust, scalable, and adaptable algorithms capable of meeting the diverse challenges of real-world applications. 

In conclusion, our research advances the understanding of MR-MTT systems and provides a framework for analyzing and optimizing the performance of distributed MRS. By leveraging the power of dimensionless variables and robust fitting methods, system designers and operators can better anticipate the behavior of MRS and make informed decisions to enhance their effectiveness in various operational contexts.


\bibliographystyle{IEEEtran}
\bibliography{bib/refs,bib/Dames}

\end{document}